\documentclass[conference]{IEEEtran}
\usepackage{units}
\usepackage{amsmath,graphicx,amssymb}
\usepackage{multirow}
\usepackage{tabulary}
\usepackage{booktabs}
\usepackage{color}
\usepackage[table]{xcolor}

\usepackage{tikz}
\usetikzlibrary{positioning,calc}
\usetikzlibrary{shapes.geometric, arrows, decorations.pathreplacing}

\def\M{{\cal M}}
\def\G{{\cal G}}

\newcommand{\bb}[1]{\textbf{#1}}

\ifCLASSINFOpdf
\else
\fi
\begin{document}
%
\title{Reliability-based Mesh-to-Grid Image Reconstruction}

\author{\IEEEauthorblockN{J\'{a}n Koloda, J\"{u}rgen Seiler and Andr\'{e} Kaup}
\IEEEauthorblockA{Chair of Multimedia Communications and Signal Processing\\
Friedrich-Alexander University (FAU) of Erlangen-Nuremberg\\
Cauerstr. 7,  91058 Erlangen, Germany\\
Email: jan.koloda@fau.de}}


%


\maketitle

\begin{abstract}
This paper presents a novel method for the reconstruction of images from samples located at non-integer positions, called mesh. This is a common scenario for many image processing applications, such as super-resolution, warping or virtual view generation in multi-camera systems. The proposed method relies on a set of initial estimates that are later refined by a new reliability-based content-adaptive framework that employs denoising in order to reduce the reconstruction error. The reliability of the initial estimate is computed so stronger denoising is applied to less reliable estimates. The proposed technique can improve the reconstruction quality by more than 2 dB (in terms of PSNR) with respect to the initial estimate and it outperforms the state-of-the-art denoising-based refinement by up to 0.7 dB.
\end{abstract}


%
\IEEEpeerreviewmaketitle

\section{Introduction}
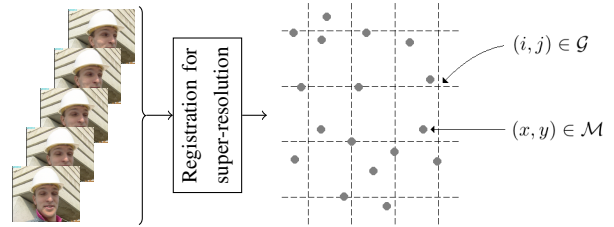
\begin{figure}[t]  
  \begin{center}
    \begin{tikzpicture}
	\node (a) {\includegraphics[width=0.1\linewidth]{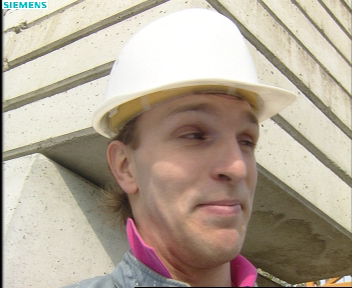}};
	\node (b) [below left = -0.45cm and -0.95cm of a] {\includegraphics[width=0.1\linewidth]{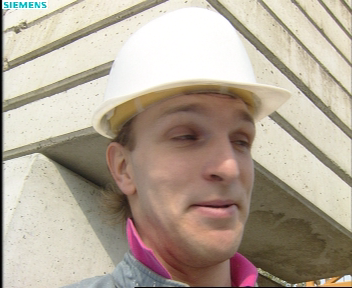}};
	\node (c) [below left = -0.45cm and -0.95cm of b] {\includegraphics[width=0.1\linewidth]{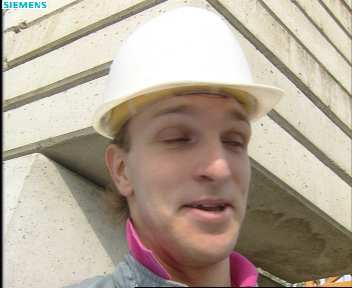}};
	\node (d) [below left = -0.45cm and -0.95cm of c] {\includegraphics[width=0.1\linewidth]{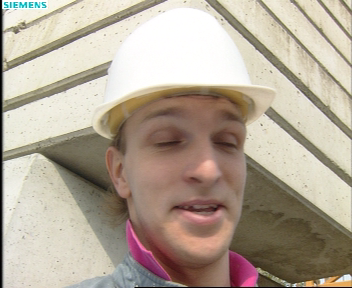}};
	\node (e) [below left = -0.45cm and -0.95cm of d] {\includegraphics[width=0.1\linewidth]{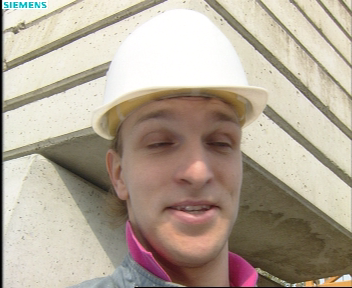}};
	
	\draw[decoration={brace},decorate] (0.5, 0.4) -- (0.5,-2.5);
	
	\node (reg) [draw=black, right=0.75cm of c,rotate=90,anchor=north,align=center] {\footnotesize{Registration for} \\ \footnotesize{super-resolution}};
	
	\node (mesh) [right = 2cm of c] {\includegraphics[width=0.5\linewidth]{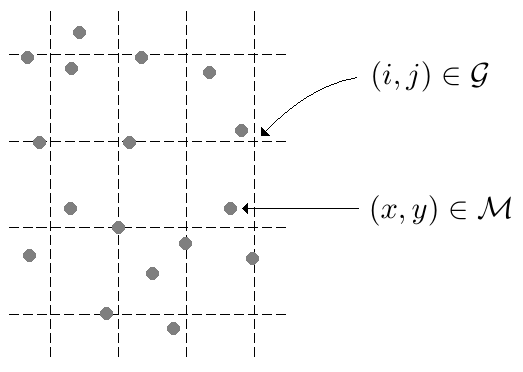}};
	
	\node (anchor) [left = 1.6cm of mesh] {};
	\draw [->] (anchor) -- (reg);
	\draw [->] (reg) -- (mesh);
	
    \end{tikzpicture}
  \end{center}  
  \caption{\small{Example of application that yields pixels at non-integer positions. The regular grid $\G$, represented by dashed lines, is addressed by integers $(i,j)$. The floating mesh $\M$ is comprised of available samples (grey dots) at non-integer positions $(x,y)$.}}
  \label{fig:definitions}  
\end{figure}

The reconstruction of images from samples located at non-integer positions is an intrinsic step in many common image processing tasks. Such is the case of frame rate up-conversion \cite{FRUC}, depth-based image rendering \cite{Fehn2004}, optical cluster eye \cite{ClusterEye} or fisheye distortion correction \cite{Fisheye}, among many others. All these applications have to deal with pixels located at non-integer positions. In the following, we will refer to these positions as floating mesh. Another example of such an application is super-resolution \cite{Superresolution}, as illustrated in Fig. \ref{fig:definitions}. The pixels from low-resolution input images are registered with sub-pixel shifts onto the high-resolution floating mesh. These registered samples at non-integer positions cannot be directly displayed nor efficiently coded so pixels on the regular grid have to be estimated. Thus, we are facing a reconstruction problem where the unknown pixels on the regular grid are estimated using the available pixels on the floating mesh.

The vast majority of image reconstruction techniques assumes that the input image is already a regular 2D array.
This is the case of backward reconstruction methods \cite{Thevenaz2000,Blu2004} that consider that the available samples lie on the regular grid and pixels on the floating mesh are to be estimated. This inverse approach can be used for invertible pixel mappings such as rotation or zoom. However, it is not applicable in cases where the available samples come from different sources or the mapping is not invertible, as is the case of the aforementioned applications. Nevertheless, there are various reconstruction algorithms that are able to deal with pixels at non-integer positions directly. The most popular methods are based on Delaunay triangulation \cite{Tesselations} followed by different types of interpolations \cite{Griddata}. More complex techniques rely on gaussian kernels \cite{Takeda2007} or distance-based filtering \cite{IDW}. Furthermore, the missing image signal can be also reconstructed using splines \cite{Lee1997}. Although the aforementioned reconstruction algorithms tend to produce acceptable results, visual artefacts, such as blurring or zig-zagging, are usually present.

In our previous work, we have developed a reconstruction framework that employs denoising as a tool to reduce the mesh-to-grid reconstruction error \cite{DBR}. This framework assumes a generic initial estimate that is refined by applying an adaptive denoising approach. Although this approach can significantly improve the reconstruction quality, it still suffers from over-smoothing and not all of the reconstruction artefacts are satisfactorily removed. In this paper, we propose a novel reliability parameter that will be employed for controlling the denoising strength. This reliability is estimated using not only the spatial configuration of the available samples but also the visual properties of the surrounding area. We design a reliability-based denoising strength controlling mechanism that aims at maximising the reconstruction quality.

The paper is organised as follows. Section \ref{sec:overview} provides an overview of the denoising-based reconstruction framework. The proposed technique is detailed in Section \ref{sec:refinement}. Simulation results are discussed in Section \ref{sec:results}. The last section is devoted to conclusions.

\section{Overview of the denoising-based reconstruction}
\label{sec:overview}
In this section, the denoising-based reconstruction (DBR) framework \cite{DBR} is briefly summarized.
Let $\G$ denote the regular grid that represents the desired image and let $\M$ be the floating mesh that contains available samples at non-integer positions, as shown in Fig. \ref{fig:definitions}. The unknown pixels in $\G$ are reconstructed using the information provided by the available samples in $\M$.
It is assumed that the image on the regular grid is estimated by a generic reconstruction algorithm. Since reconstruction errors are present, the estimated image $\tilde{I}$ can be regarded as a noisy version of the original image $I$, i.e.,
\begin{equation}
	\tilde{I}(i,j) = I(i,j) + \eta(i,j) \quad \forall (i,j) \in \G .
	\label{noisy_pixel}
\end{equation}
This suggests that a denoising procedure can be applied in order to enhance the reconstruction quality. Note that cancelling the noise $\eta$ is equivalent to reducing the reconstruction error. 

The objective of DBR is to set the denoising strength according to the reconstruction error, so high reconstruction errors will be denoised strongly while small reconstruction errors will be denoised only slightly or not at all. Since the true reconstruction error cannot be computed, it has been proposed in \cite{DBR} to use the amount of effective data around the missing pixel as an indicator of the expected reconstruction error. This amount, $E$, is computed by placing an exponentially decaying isotropic window centred at the missing pixel and summing the contribution of the surrounding available samples located on the floating mesh, i.e.,
\begin{equation}
	E(i,j) = \sum_{(x,y) \in \M} \text{e}^{-\sqrt{(i-x)^2 + (j-y)^2}} \quad \forall (i,j) \in \G.
	\label{effective_data_old}
\end{equation}
Note that unlike the true reconstruction error, the amount of effective data can be computed without any knowledge about the reference image.

Finally, it has been shown in \cite{DBR} that the denoising strength can be linked to the amount of the effective data by means of a sigmoid shaped function that requires training a set of three parameters. Although the DBR framework is applicable to any denoising algorithm, in \cite{DBR} it has been coupled to the block-matching and 3D filtering (BM3D) algorithm \cite{BM3D} since it is one of the most efficient denoising methods \cite{NonGaussianNoise}.

\section{Reliability-based reconstruction}
\label{sec:refinement}
In this section, we introduce a novel approach to control the denoising strength based on pixel reliability. By doing so, we not only take into account the spatial layout of the available samples but also the visual properties of the surrounding area. We first derive a content-adaptive method of estimating the reliability of the initial estimate $\tilde{I}(i,j)$. Second, we develop a procedure to employ the estimated reliability to control the denoising strength in order to improve the reconstruction quality. Both issues are addressed in the next subsections.

\subsection{Reliability of the initial estimate}
DBR uses the amount of effective data as an indicator of the expected reconstruction error so the denoising strength can be set accordingly. Although DBR can significantly improve the resulting quality, it still suffers from two main drawbacks:
\begin{enumerate}
 \item To compute the amount of effective data (see (\ref{effective_data_old})) all samples, surrounding the pixel to be reconstructed, are taken into account regardless of their correlation with the missing pixel or whether they have been used for estimating it. This is especially inconvenient for triangulation-based techniques where only the three available samples of the corresponding triangle are used for reconstruction no matter how close the rest of the available samples is.
 \item DBR controls the denoising strength only according to the spatial layout of the available samples. However, complex image structures, such as edges or textures, are more prone to yield higher reconstruction errors and, therefore, should be subjected to stronger denoising.
\end{enumerate}

In order to reduce the contribution of uncorrelated samples, we propose to limit the computation of the effective data to the involving triangle. Let $\triangle_{i,j}$ denote the Delaunay triangle \cite{Tesselations} enclosing the pixel at position $(i,j)$ of the regular grid so the corresponding amount of effective data is computed as
\begin{equation}
	E_{\triangle}(i,j) = \sum_{(x,y) \in \triangle_{i,j}} \text{e}^{-\sqrt{(i-x)^2 + (j-y)^2}} \quad \forall (i,j) \in \G.
	\label{effective_data}
\end{equation}
Note that the triangle is the smallest convex structure surrounding a pixel. Therefore, and given the high spatial correlation of natural images \cite{MathAnalysisImages}, the three available samples comprising the triangle are highly relevant for computing the initial estimate even for non-triangulation-based techniques.

\begin{figure}[t]	
	\centering 
	\includegraphics[width=1.0\linewidth]{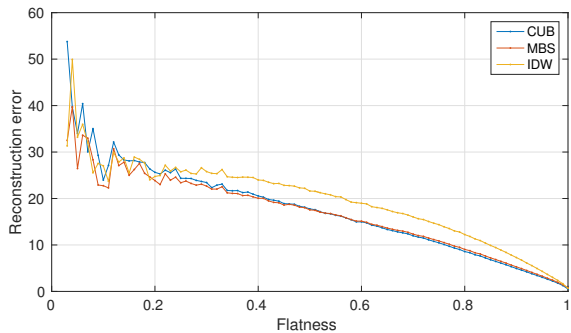}
	\caption{\small{Average reconstruction error as a function of visual flatness for different reconstruction algorithms: cubic interpolation (CUB) \cite{Griddata}, multilevel B-splines (MBS) \cite{Lee1997} and inverse distance weighting (IDW) \cite{IDW}. Tecnick image dataset is employed.}}
	\label{fig:flatness}
\end{figure}

On the other hand, complex structures and fine details are generally difficult to accurately reconstruct. Therefore, it is to expect larger reconstruction errors in visually heterogeneous regions. In order to detect whether a pixel belongs to such a region, we propose to use the visual flatness parameter \cite{Koloda2016}. The visual flatness, $F$, of a pixel at position $(i,j)$ is related to the dynamic range of the available samples comprising the corresponding Delaunay triangle $\triangle(i,j)$, i.e.,
\begin{multline}
	F(i,j) = 1 - \Big( \max \left( S(x,y | x,y \in \triangle_{i,j}) \right) \\ -  \min \left( S(x,y | x,y \in \triangle_{i,j}) \right) \Big) \quad \forall (i,j) \in \G.
	\label{flatness}
\end{multline}
where $S(x,y) \in [0, 1]$ is the value of the sample located at $(x,y) \in \M$. Completely homogeneous regions yield maximum flatness (equal to 1) and it decreases to zero for strong edges and highly textured areas. Figure \ref{fig:flatness} shows the average reconstruction error as a function of visual flatness for different reconstruction methods using the Tecnick image dataset \cite{Tecnick}. It follows that the visual flatness is a good indicator of the expected reconstruction quality and it can be computed using only the available samples from the floating mesh.

Finally, we define the reliability of the initial pixel estimate at position $(i,j) \in \G$ as a weighted combination of the surrounding amount of effective data and the corresponding visual flatness, i.e.,
\begin{equation}
	R_{\lambda}(i,j) = (1 - \lambda) E_{\triangle}(i,j) + \lambda F(i,j)
	\label{reliability}
\end{equation}
where $\lambda \in [0, 1]$ is the balance parameter that controls the contribution of the effective data and the visual flatness, respectively. It follows that highly reliable estimates are achieved for pixels belonging to homogeneous regions and tightly enclosed by the three available samples comprising the corresponding Delaunay triangle. On the other hand, the estimates are regarded as not reliable if they are enclosed by large triangles with disparate sample values.

\subsection{Adaptive denoising strength selection}
\label{subsec:strength_selection}

\begin{figure}[t]	
	\centering
	\includegraphics[width=1.0\linewidth]{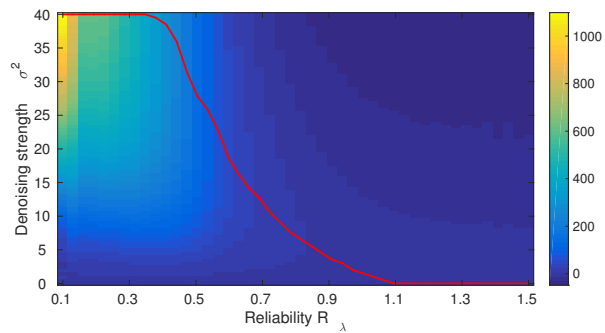}
	\caption{\small{Average gain per pixel for different values of $R_{\lambda}$ and $\sigma^2$. In this example, the balance parameter $\lambda$ is set to 0.5. The path of maximum accumulated gain is depicted by red colour. Tecnick image dataset is employed.}}
	\label{fig:gain_chart}
\end{figure}

As shown in the previous section, the reliability of an initial estimate is a suitable indicator of the expected reconstruction error. Therefore, in order to achieve high quality reconstructions, we propose to use the reliability parameter to adjust the denoising strength. In order to do so, we define the gain, $G$, of the denoised image with respect to the initial estimate \cite{DBR},
\begin{equation}
	\begin{array}{l l l}
		G & = & \varepsilon(i,j) - \varepsilon_D(i,j) \\
		  & = & \left( I(i,j) - \tilde{I}(i,j) \right)^2 - \left( I(i,j) - \tilde{I}_{\sigma^2}(i,j) \right)^2 \\
		 & & \forall (i,j) \in \G.
	\end{array}
\label{gain}
\end{equation}
where $\tilde{I}_{\sigma^2}$ represents the denoised version of the initial estimate $\tilde{I}$. The denoising is carried out by BM3D using $\sigma^2$ as the noise power parameter. Based on (\ref{gain}), we define the average gain per pixel $\bar{G}_{\lambda}(\sigma^2, R_{\lambda})$ as a function of $\sigma^2$ and $R_{\lambda}$. Figure \ref{fig:gain_chart} shows $\bar{G}_{\lambda}(\sigma^2, R_{\lambda})$ for $\lambda = 0.5$. It can be observed that large gains are achieved when low reliability estimates are denoised strongly. On the other hand, high reliability estimates should be denoised only slightly, otherwise no gain or even quality loss can occur. In order to achieve the best quality reconstruction, the path of the maximum accumulated gain $G_A$ has to be followed, where
\begin{equation}
	G_A(\lambda) = \displaystyle{\int \max \left( \bar{G}_{\lambda} (\sigma^2, R_{\lambda}) \right) \text{d}R_{\lambda}}.
	\label{maximum_accumulated_gain}
\end{equation}
This path is also indicated in Fig. \ref{fig:gain_chart}. It follows that for a given $\lambda$ the curve of the maximum gain can be well modelled by a clipped exponential function. Thus, in order to link the best fitting denoising strength $\hat{\sigma}^2$ with the pixel reliability we propose to use the following exponential approximation

\begin{equation}
	\hat{\sigma}^2(i,j) = \max \left(0, \; \min \left( \alpha \exp(\beta R_{\lambda}(i,j)), \; \sigma^2_{max} \right) \right)
	\label{sigma}
\end{equation}
where $\hat{\sigma}^2$ saturates to zero for negative outputs and positive outputs are clipped by an upper boundary $\sigma^2_{max}$ that is set to 40 as in \cite{DBR}. It follows that $\hat{\sigma}^2 = \hat{\sigma}^2(\alpha,\beta,\lambda)$ is a function of three parameters. These parameters can be estimated by optimising the reconstruction performance. In order to do so, we propose to maximise the expected accumulated gain $G_E$, which we define as follows
\begin{equation}
	G_E(\alpha, \beta, \lambda) = \displaystyle{\int \bar{G}_{\lambda}\left(\hat{\sigma}^2\left(\alpha, \beta, \lambda \right), R_{\lambda} \right) p(R_{\lambda}) \text{d}R_{\lambda}}
	\label{accumulated_gain}
\end{equation}
where $p(R_{\lambda})$ is the probability of a pixel having the reliability equal to $R_{\lambda}$. Note that from the modelling point of view the gains associated to estimates with uncommon reliabilities are neglected and effectively treated as outliers.
The optimal values for the three parameters are obtained by maximising the expected accumulated gain, i.e.,
\begin{equation}
	(\alpha', \beta', \lambda') = \underset{\alpha, \beta, \lambda}{\text{argmax}} \left( G_E(\alpha, \beta, \lambda) \right).
	\label{optimal_parameters}
\end{equation}
The estimated values are detailed in Section \ref{sec:results}. Finally, the proposed reliability-based mesh-to-grid reconstruction algorithm (RMG) is summarized as follows:
\begin{enumerate}
	\item Obtain an initial image estimate by applying a suitable reconstruction technique.	
	\item For every initial estimate on the regular grid, compute the corresponding reliability $R_{\lambda}$ using (\ref{reliability}).	
	\item Use $R_{\lambda}$ to estimate the best fitting denoising strength $\hat{\sigma}^2$ according to (\ref{sigma}).	
	\item Finally, for every pixel on the regular grid apply BM3D with the noise power parameter set to the corresponding value of $\hat{\sigma}^2$.
\end{enumerate}

\section{Simulation Results}
\label{sec:results}

\begin{table}[b]
	\centering
	\small
	\begin{tabular*}{0.48\textwidth}{l| @{\extracolsep{\fill}} ccccccc}
		\toprule
				 & LIN    & CUB    & NNI    & NNB    & IDW    & BSR    & KER  	\\ \hline
		$\alpha'$        & 214.0  & 298.0  & 185.0  & 133.0  & 216.0  & 318.0  & 394.0  \\
		$\beta'$         & -4.3   & -4.5   & -4.4   & -2.5   & -3.5   & -4.7   & -4.8	\\
		$\lambda'$       & 0.6    & 0.6    & 0.6    & 0.9    & 0.5    & 0.3    & 0.2 	\\
		\bottomrule
	\end{tabular*}
	\vspace{0.25cm}
	\caption{\small{Estimated parameters of $\alpha$, $\beta$ and $\lambda$ for different initial reconstruction algorithms. Tecnick image dataset is employed.}}
	\label{tab:parameters}
\end{table}

\begin{figure}[t]  
  \begin{center}
      \begin{tikzpicture}
	\node (a) {\includegraphics[width=0.47\linewidth]{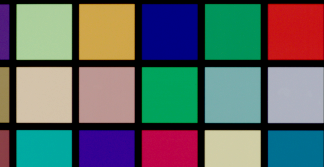}};
	\node (b) [right = 0.05cm of a] {\includegraphics[width=0.47\linewidth]{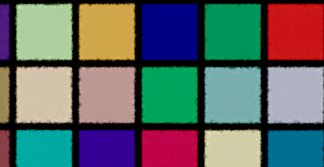}};
	\node (c) [below = 0.3cm of a] {\includegraphics[width=0.47\linewidth]{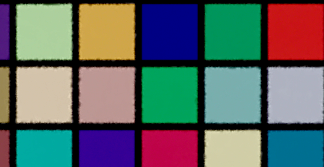}};	
	\node (d) [right = 0.05cm of c] {\includegraphics[width=0.47\linewidth]{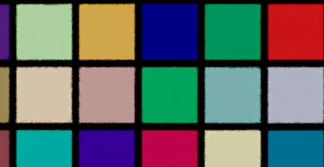}};
	
	\node (a_label) [below = -0.15cm of a] {\small{(a)}};
	\node (b_label) [below = -0.15cm of b] {\small{(b)}};
	\node (c_label) [below = -0.15cm of c] {\small{(c)}};
	\node (d_label) [below = -0.15cm of d] {\small{(d)}};
	
	\node (b_psnr) [above right = -0.55cm and -1.3cm of b] {\footnotesize{ \color{white}{30.85dB}}};
	\node (c_psnr) [above right = -0.55cm and -1.3cm of c] {\footnotesize{ \color{white}{31.68dB}}};
	\node (d_psnr) [above right = -0.55cm and -1.3cm of d] {\footnotesize{ \color{white}{33.51dB}}};
      \end{tikzpicture}
  \end{center}  
  \caption{\small{Subjective comparison of the reconstruction quality using a sample ratio of 50\%. (a) Original image. (b) Initial estimation by MBS. (c) Reconstructed by DBR. (d) Reconstructed by RMG. The corresponding PSNR values (in dB) are also indicated. (Best viewed enlarged on a screen.)}}
  \label{fig:color}
\end{figure}
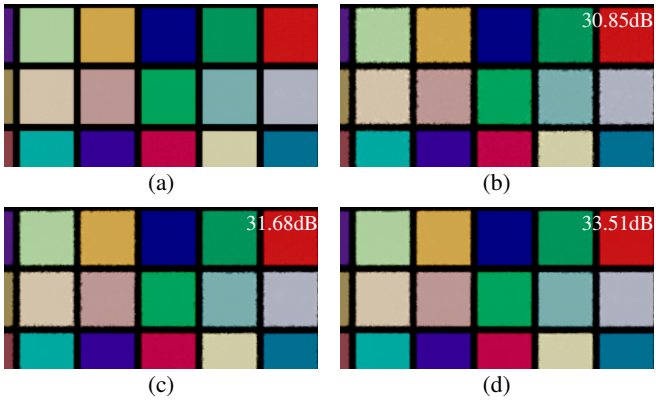 

In order to test the performance, both the reference image (i.e., the regular grid) and the available floating mesh need to be known. However, since to the best of our knowledge, there is no such ground-truth image database available, we have designed in \cite{DBR} a reconstruction framework that consists of utilising a random subset of available samples located at non-integer positions. By doing so, we cover a large variety of applications, such as super-resolution or image rendering. This framework is detailed in \cite{DBR} and we briefly revisit it here. We employ the ARRI image dataset \cite{ARRI} of 11 images with dimensions of 2880$\times$1620 whose pixels are addressed by a pair of integers $(m,n)$. These images are filtered by a low-pass antialiasing filter with the digital cut-off frequency of 1/$\phi$ where $\phi = 5$ as in \cite{DBR}. The floating mesh of pixels at non-integer positions is simulated by assuming that $(x,y) = \left( \nicefrac{m}{\phi}, \nicefrac{n}{\phi} \right)$, where $(x,y) \in \M$. In the same way, the regular grid (i.e. the reference image) is comprised by all the pixels at positions $(i,j) = \left( \nicefrac{m}{\phi}, \nicefrac{n}{\phi} \right)$ such that $(m,n)$ are both multiples of $\phi$ so $(i,j) \in \G$ is a pair of integers. Finally, the pixels from the floating mesh $\M$ are randomly sampled and they are used for reconstructing the pixels on the regular grid $\G$. For more details please refer to \cite{DBR}.

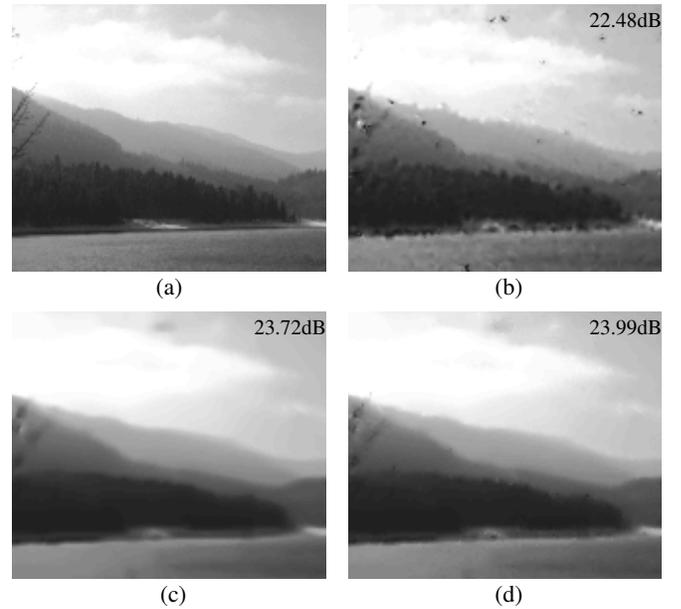
\begin{figure}[t]  
  \begin{center}
      \begin{tikzpicture}
	\node (a) {\includegraphics[width=0.47\linewidth]{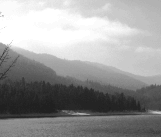}};
	\node (b) [right = 0.05cm of a] {\includegraphics[width=0.47\linewidth]{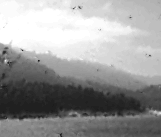}};
	\node (c) [below = 0.3cm of a] {\includegraphics[width=0.47\linewidth]{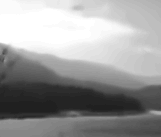}};
	\node (d) [right = 0.05cm of c] {\includegraphics[width=0.47\linewidth]{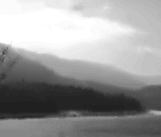}};
	
	\node (a_label) [below = -0.15cm of a] {\small{(a)}};
	\node (b_label) [below = -0.15cm of b] {\small{ (b)}};
	\node (c_label) [below = -0.15cm of c] {\small{ (c)}};
	\node (d_label) [below = -0.15cm of d] {\small{ (d)}};
	
	\node (b_psnr) [above right = -0.55cm and -1.3cm of b] {\footnotesize{ \color{black}{22.48dB}}};
	\node (c_psnr) [above right = -0.55cm and -1.3cm of c] {\footnotesize{ \color{black}{23.72dB}}};
	\node (d_psnr) [above right = -0.55cm and -1.3cm of d] {\footnotesize{ \color{black}{23.99dB}}};
      \end{tikzpicture}
  \end{center}   
  \caption{\small{Subjective comparison of the reconstruction quality using a sample ratio of 30\%. (a) Original image. (b) Initial estimation by KER. (c) Reconstructed by DBR. (d) Reconstructed by RMG. The corresponding PSNR values (in dB) are also indicated. (Best viewed enlarged on a screen.)}}
  \label{fig:lake}  
\end{figure}

Sample ratios from 20\% up to 80\% are used. Sample ratio denotes the ratio between the number of available samples and the total amount of pixels an image is comprised of. The initial estimates are computed using linear interpolation (LIN) \cite{Griddata}, cubic interpolation (CUB) \cite{Griddata},
natural neighbour interpolation (NNI) \cite{NNI}, nearest neighbour approach (NNB) \cite{Griddata}, inverse distance weighting (IDW) \cite{IDW}, multi-level B-spline
reconstruction (MBS) \cite{Lee1997} and kernel-based reconstruction (KER) \cite{Takeda2007}. The parameters $\alpha$, $\beta$ and $\lambda$ are estimated according to (\ref{optimal_parameters}) using the Tecnick database and are summarized in Table \ref{tab:parameters}.
Note that the Tecnick database is independent from the ARRI dataset which is used to test the performance.

\begin{table}[t]
	\centering
	\small
	\begin{tabulary}{0.95\linewidth}{LC|CCCCCCC}
		\toprule
		\multicolumn{1}{p{0.1cm}}{} & \multicolumn{1}{p{0.1cm}}{}  & \multicolumn{7}{c}{Sample ratio} \\
					&	& 20\%       & 30\%       & 40\%       & 50\%       & 60\%       & 70\%       & 80\% 	 	\\ \hline
		\multirow{2}{*}{\parbox{0.4cm}{LIN}} 	& \multicolumn{1}{p{0.2cm}|}{(a)}	& 0.25       & 0.32       & 0.30       & 0.20       & 0.11       & 0.05       & 0.02  	 	\\
							& \multicolumn{1}{p{0.2cm}|}{(b)}	& \bb{0.44}  & \bb{0.49}  & \bb{0.51}  & \bb{0.50}  & \bb{0.47}  & \bb{0.44}  & \bb{0.41}	\\ \hline
					
		\multirow{2}{*}{\parbox{0.4cm}{CUB}}    & \multicolumn{1}{p{0.2cm}|}{(a)}	& 0.43       & 0.46       & 0.46       & 0.38       & 0.27       & 0.17       & 0.09 	 	\\
							& \multicolumn{1}{p{0.2cm}|}{(b)}	& \bb{0.62}  & \bb{0.67}  & \bb{0.68}  & \bb{0.68}  & \bb{0.64}  & \bb{0.62}  & \bb{0.58}	\\ \hline
					
		\multirow{2}{*}{\parbox{0.4cm}{NNI}}    & \multicolumn{1}{p{0.2cm}|}{(a)}	& 0.14       & 0.17       & 0.09       & 0.03       & 0.01       & 0.01       & 0.00  	 	\\
							& \multicolumn{1}{p{0.2cm}|}{(b)}	& \bb{0.21}  & \bb{0.27}  & \bb{0.30}  & \bb{0.30}  & \bb{0.29}  & \bb{0.27}  & \bb{0.24}  	\\ \hline
					
		\multirow{2}{*}{\parbox{0.4cm}{NNB}}    & \multicolumn{1}{p{0.2cm}|}{(a)}	& 1.82       & 1.78       & 1.72       & 1.66       & 1.58       & 1.50       & 1.42  	 	\\
							& \multicolumn{1}{p{0.2cm}|}{(b)}	& \bb{2.12}  & \bb{2.21}  & \bb{2.24}  & \bb{2.24}  & \bb{2.21}  & \bb{2.18}  & \bb{2.16}  	\\ \hline
	
		\multirow{2}{*}{\parbox{0.4cm}{IDW}}    & \multicolumn{1}{p{0.2cm}|}{(a)}	& 0.36       & 0.27       & 0.26       & 0.27       & 0.29       & 0.29       & 0.29  	 	\\
							& \multicolumn{1}{p{0.2cm}|}{(b)}	& \bb{0.62}  & \bb{0.62}  & \bb{0.64}  & \bb{0.63}  & \bb{0.63}  & \bb{0.61}  & \bb{0.58}  	\\ \hline
					
		\multirow{2}{*}{\parbox{0.4cm}{MBS}}    & \multicolumn{1}{p{0.2cm}|}{(a)}	& 0.21       & 0.29       & 0.28       & 0.19       & 0.10       & 0.04       & 0.01  	 	\\
							& \multicolumn{1}{p{0.2cm}|}{(b)}	& \bb{0.35}  & \bb{0.41}  & \bb{0.44}  & \bb{0.45}  & \bb{0.45}  & \bb{0.44}  & \bb{0.41}  	\\ \hline
					
		\multirow{2}{*}{\parbox{0.4cm}{KER}}    & \multicolumn{1}{p{0.2cm}|}{(a)}	& 1.31       & 0.73       & 0.46       & 0.27       & 0.14       & 0.08       & 0.04  	 	\\
							& \multicolumn{1}{p{0.2cm}|}{(b)}	& \bb{1.64}  & \bb{1.00}  & \bb{0.71}  & \bb{0.48}  & \bb{0.32}  & \bb{0.23}  & \bb{0.15}  	\\ \bottomrule
					
	\end{tabulary}
	\vspace{0.25cm}
	\caption{\small{Average PSNR gains (in dB) for (a) DBR and (b) the proposed RMG. Different reconstruction techniques are used as initial estimates and different sample ratios are employed. The best performances are in bold face. ARRI image dataset is employed.}}
	\label{tab:results}	
\end{table}

Since the proposed technique can be considered as a generic posterior refinement procedure over an initial estimate, it makes sense to evaluate the performance in terms of gain. Table \ref{tab:results} shows the average PSNR gain (in dB) of DBR and the proposed RMG with respect to the tested reconstruction techniques. It is observed that the proposed technique can improve the average reconstruction quality by more than 2 dB, depending on the initial estimate. Moreover, average gains of up to 0.7 dB with respect to DBR can be achieved. This improvement is also seen at subjective level, as illustrated in Figs. \ref{fig:color} and \ref{fig:lake}. It is shown that RMG yields sharper results, better defined details and it is able to successfully remove most of the reconstruction artefacts. Finally, note that for higher sample ratios RMG provides considerable gains while DBR gains drop almost to zero in most of the cases.

\section{Conclusion}
In this paper, we consider the problem of image reconstruction from pixels located at non-integer positions, called mesh.
The proposed method assumes that the image on the regular grid is first reconstructed by a generic reconstruction algorithm.
This initial estimate is later refined by a denoising-based procedure that adaptively adjusts the denoising strength according to the reliability of the initial estimate. This reliability takes into account the spatial layout of the available samples as well as the visual properties of the surrounding area. Simulations reveal that PSNR gains of more than \mbox{2 dB} with respect to the initial estimate can be achieved. In addition, our proposal clearly outperforms previous denoising-based refinement both on objective and subjective level.

Ongoing work is focused on testing the proposed algorithm on various image processing applications, e.g., super-resolution.


\section*{Acknowledgment}
This work has been partially supported by the Research Training Group 1773 ''Heterogeneous Image Systems'', funded by the German Research Foundation (DFG).



%

\bibliographystyle{IEEEbib}
\bibliography{refs}

\end{document}